# Algorithm for Training Neural Networks on Resistive Device Arrays

**Authors:**  Tayfun Gokmen* and Wilfried Haensch

**Affiliations**
IBM Research AI, Yorktown Heights, NY 10598 USA
*Correspondence to:  tgokmen@us.ibm.com

**Abstract**
Hardware architectures composed of resistive cross-point device arrays can provide significant power and speed benefits for deep neural network training workloads using stochastic gradient descent (SGD) and backpropagation (BP) algorithm. The training accuracy on this imminent analog hardware however strongly depends on the switching characteristics of the cross-point elements. One of the key requirements is that these resistive devices must change conductance in a symmetrical fashion when subjected to positive or negative pulse stimuli. Here, we present a new training algorithm, so-called the "Tiki-Taka" algorithm, that eliminates this stringent symmetry requirement. We show that device asymmetry introduces an unintentional implicit cost term into the SGD algorithm, whereas in the "Tiki-Taka" algorithm a coupled dynamical system simultaneously minimizes the original objective function of the neural network and the unintentional cost term due to device asymmetry in a self-consistent fashion. We tested the validity of this new algorithm on a range of network architectures such as fully connected, convolutional and LSTM networks. Simulation results on these various networks show that whatever accuracy is achieved using the conventional SGD algorithm with symmetric (ideal) device switching characteristics the same accuracy is also achieved using the "Tiki-Taka" algorithm with non-symmetric (non-ideal) device switching characteristics. Moreover, all the operations performed on the arrays are still parallel and therefore the implementation cost of this new algorithm on array architectures is minimal; and it maintains the aforementioned power and speed benefits. These algorithmic improvements are crucial to relax the material specification and to realize technologically viable resistive crossbar arrays that outperform digital accelerators for similar training tasks.

# INTRODUCTION

In the past few years, deep neural networks (DNN) [1] have made tremendous advances, in some cases surpassing human level performance, tackling challenging problems such as speech recognition [2] [3], natural language processing [4] [5], image classification [6] [7] [8], and machine translation [9]. Training of large DNNs, however, is a time consuming and computationally intensive task that demands datacenter scale computational resources composed of state of the art GPUs [6] [10]. There have been many attempts to accelerate deep learning workloads beyond GPUs by designing custom hardware utilizing reduced precision arithmetic to improve the throughput and energy efficiency of the underlying CMOS technology [11]. Alternative to digital approaches, resistive cross-point device arrays are proposed to further increase the throughput and energy efficiency of the overall system by performing the vector-matrix multiplications in the analog domain. In addition, these device arrays can perform the weight update operation locally with no weight movement and therefore they bring further benefits compared to digital approaches.

Resistive cross-point devices, so called resistive processing unit (RPU) [12] device arrays that can simultaneously store and process data locally and in parallel, are promising candidates for intensive DNN training workloads. The concept of using resistive cross-point device arrays [12] [13] [14] [15] [16] [17] [18] as DNN accelerators has been tested on a variety of network architectures and datasets mainly by simulations but also with some limited hardware demonstrations. Considering state-of-the-art learning algorithms, to par training accuracy compared to the conventional digital hardware a restrictive set of RPU device specifications must be met. As shown empirically [12] [19] [20], a key requirement is that these analog resistive devices must change conductance symmetrically when subjected to positive or negative voltage pulse stimuli. This requirement differs significantly from those needed for memory elements and accomplishing such symmetrically switching analog devices is a difficult task. Substantial efforts are devoted to engineer new material stacks or adopt the existing ones, originally developed for memory [17] and battery [15] [21] applications, to achieve the symmetry criteria needed for DNN training. Besides material engineering efforts, CMOS only [22] and CMOS assisted solutions in tandem with existing memory device technologies [23] are also considered but introduce an overhead of making the cross-point element increasingly more complex.

Here we present a new technique that can address the issue of non-symmetric device switching characteristics at the algorithm level, in a physical-hardware invariant form. In the rest of the paper we show that the device switching characteristics introduces an additional cost term into the optimization objective of the conventional SGD algorithm. The presence of this additional term entails poor training results for non-symmetric devices as the system is in competition with minimizing the objection function of the neural network against this unintentional cost term. In this new technique we introduce a coupled dynamical system that simultaneously minimizes the objective function of the original SGD algorithm as well as the unintentional cost term due to device asymmetry in a self-consistent fashion. This algorithm learns by continuously exchanging information between two system's components and hence we call it the "Tiki-Taka" algorithm. We show that the "Tiki-Taka" algorithm is general enough to handle a large range of asymmetric device switching behaviors and therefore applicable to a variety of device technologies. We tested the algorithm by performing training simulations using various device switching characteristics on three different network architectures: fully connected, convolutional and LSTMs. In all cases the results of the training performed with the "Tiki-Taka" algorithm using non-ideal device characteristics are indistinguishable from the ones achieved with the SGD algorithm using ideal devices. We also discuss the implementation cost of the "Tiki-Taka" algorithm on realistic RPU device arrays in terms of area, power and speed and show that the overall cost is minimal.

## MATERIALS AND METHODS

### Array Operations: Forward, Backward and Update

The stochastic gradient descent (SGD) using the backpropagation algorithm is composed of three cycles -- forward, backward and weight update -- that are repeated many times until a convergence criterion is met. For a single fully connected layer where $N$ inputs neurons are connected to $M$ output (or hidden) neurons, the forward cycle involve computing a vector-matrix multiplication ($\boldsymbol{y} = \boldsymbol{W}\boldsymbol{x}$) where the vector $\boldsymbol{x}$ of length $N$ represents the activities of the input neurons and the matrix $\boldsymbol{W}$ of size $M \times N$ that stores the weight values between each pair of input and output neurons. The resulting vector $\boldsymbol{y}$ of length $M$ is further processed by performing a non-linear activation on each of the elements and then passed to the next layer. Once the information reaches to the final output layer, the error signal is calculated and backpropagated through the network. The backward cycle on a single layer also involves a vector-matrix multiplication on the transpose of the weight matrix ($\boldsymbol{z} = \boldsymbol{W}^\mathsf{T}\boldsymbol{\delta}$), where the vector $\boldsymbol{\delta}$ of length $M$ represents the error calculated by the output neurons and the vector $\boldsymbol{z}$ of length $N$ is further processed using the derivative of neuron non-linearity and then passed down to the previous layers. Finally, in the update cycle the weight matrix $\boldsymbol{W}$ is updated by performing an outer product of the two vectors that are used in the forward and the backward cycles and usually expressed as $\boldsymbol{W} \leftarrow \boldsymbol{W} + \eta\,(\boldsymbol{x} \otimes \boldsymbol{\delta})$ where $\eta$ is a global learning rate. Consistently, the SGD update rule for each parameter $w_{ij}$ corresponding to $i^{\text{th}}$ column and $j^{\text{th}}$ row (the layer index is dropped for simplicity) can be written as

$$w_{ij} \leftarrow w_{ij} - \eta \Delta w_{ij} \qquad (1)$$

where $\Delta w_{ij}$ is the gradient of the objective function with respect to parameter $w_{ij}$, and has a form $\Delta w_{ij} = x_i \times \delta_j$, where $x_i$ is the input activation for the $i^{\text{th}}$ column and $\delta_j$ is the backpropagated error signal for the $j^{\text{th}}$ row.

The above three operations performed on the weight matrix $\boldsymbol{W}$ during the SGD\BP algorithm are implemented using 2D crossbar arrays of resistive devices all in parallel and constant time using the physical properties of the array. For instance, the stored conductance values in the crossbar array form the matrix $\boldsymbol{W}$, however physically only positive conductance values are allowed and therefore to encode both positive and negative weight values a pair of RPU devices is operated in differential mode. For each parameter $w_{ij}$ in the weight matrix $\boldsymbol{W}$, there exists two devices that encode a single weight value

$$w_{ij} = \mathrm{K}(g_{ij} - g_{ij,ref}) \qquad (2)$$

where $g_{ij}$ is the conductance value stored on the first RPU device, $g_{ij,ref}$ is the conductance value stored on the second device used as a reference both corresponding to $i^{th}$ column and $j^{th}$ row and K is the gain factor that is controlled by a combination of factors, such as integration time, integration capacitor and voltage levels, at the peripheral circuitry. In the forward cycle, the input vector $x$ is transmitted as voltage pulses through each of the columns and resulting vector $y$ is read as a differential current signal from the rows [24]. Similarly, the backward cycle can be performed by inputting voltage pulses from the rows and reading the results from the columns. These two cycles simply rely on Ohm's law and the Kirchhoff's law in order to perform the vector-matrix multiplications. In contrast to the forward and backward cycles, implementing the update cycle is trickier and employs the device switching characteristics to trigger the necessary conductance change $\Delta g_{ij}$ that should practically match the required weight change $\eta \Delta w_{ij}$ of the SGD algorithm, such that $K\Delta g_{ij} \cong \eta(x_i \times \delta_j)$. To perform the local multiplication operation needed to calculate $\Delta w_{ij} = x_i \times \delta_j$, different pulse encoding schemes [12] [25] [17] are proposed all of which reduce the multiplication to a simple coincidence detection that can be realized by RPU devices. For instance, in the stochastic update scheme numbers that are encoded from the columns and rows ($x_i$ and $\delta_j$) are translated to stochastic bit streams using stochastic translators [12]. Then they are sent to the crossbar array where each RPU device is expected to change its conductance by a small amount $\Delta g_{min}$ when bits from $x_i$ and $\delta_j$ coincide. In this scheme the weight update happens as a series of coincidence events each triggering a conductance increment (or decrement) and the expected number of coincidences is

$$\mathbb{E}(\#\ of\ pulse\ coincidences\ at\ i^{th}\ column\ and\ j^{th}\ row) = \frac{\eta(x_i \times \delta_j)}{K\Delta g_{min}} \quad (3)$$

where $K\Delta g_{min} \triangleq \Delta w_{min}$ is the smallest expected weight change due to a single coincidence event. We note the pulses are generated simultaneously at the array periphery assuming a knowledge of a single K value and a single $\Delta g_{min}$ value that hopefully holds true for the whole array. However, as we show next the real changes triggered by each RPU device may differ from the expected values and therefore create artifacts into the SGD algorithm.

**Expected vs. Actual Weight Update**

Using the formula of the expected number of pulse coincidences from Eq. 3, the actual algorithmic weight change caused by the update cycle performed by the RPU devices can be derived as

$$\Delta w_{ij,actual} = \Delta w_{ij} \begin{cases} \dfrac{\Delta g_{ij}^p(g_{ij})}{\Delta g_{min}} & if\ \Delta w_{ij} < 0 \\[2ex] \dfrac{\Delta g_{ij}^n(g_{ij})}{\Delta g_{min}} & if\ \Delta w_{ij} > 0 \end{cases} \quad (4)$$

where $\Delta g_{ij}^p$ and $\Delta g_{ij}^n$ are the actual device responses for the incremental conductance changes for positive and negative branches and they may be functions of the current device conductance $g_{ij}$. We assume that the update pulses are applied only to the first set of RPU devices and the reference devices are kept constant. However, to enable both positive and negative conductance changes the polarity of the pulses are switched during the update cycle and hence there exists two branches for each device used for the updates. Plugging Eq. 4 into the SGD update rule from Eq. 1 results in an actual update rule implemented by the RPU devices

$$w_{ij} \leftarrow w_{ij} - \eta \Delta w_{ij} \left[ \frac{\Delta g_{ij}^p(g_{ij}) + \Delta g_{ij}^n(g_{ij})}{2\, \Delta g_{min}} \right] - \eta |\Delta w_{ij}| \left[ \frac{\Delta g_{ij}^p(g_{ij}) - \Delta g_{ij}^n(g_{ij})}{2\, \Delta g_{min}} \right] \quad (5)$$

that captures the deviation of the expected device conductance changes from the actual ones realized by the RPU devices. Without loss of generality Eq. 5 can be rewritten as

$$w_{ij} \leftarrow w_{ij} - \eta \Delta w_{ij} F_{ij}(w_{ij}) - \eta |\Delta w_{ij}| G_{ij}(w_{ij}) \quad (6)$$

where $F_{ij}(w_{ij})$ and $G_{ij}(w_{ij})$ are the symmetric (additive) and antisymmetric (subtractive) combinations of the positive and negative update branches parametrized using the weight values corresponding $i^{th}$ column and $j^{th}$ row. Note that the functions $F_{ij}$ and $G_{ij}$ can generally be functions of the current weight value $w_{ij}$ as well as vary from one cross-point to another due to device-to-device variability. Although we used the stochastic pulsing scheme for the derivation of Eq. 5 and 6, the equations are general and do not depend on the underlying pulse implementations. Table 1 compares the desired SGD update rule (Eq. 1) to the hardware induced update rule (Eq. 6), that has contributions from the device switching characteristics.

**Table 1. Summary of the Update Rules**

| *Desired SGD Update Rule* | | *Hardware Induced Update Rule* | |
|---|---|---|---|
| $w \leftarrow w - \eta \Delta w$ | (7) | $w \leftarrow w - \eta \Delta w F(w) - \eta |\Delta w| G(w)$ | (8) |

All sub-indexes corresponding to $i^{th}$ column, $j^{th}$ row and the layer index is dropped for simplicity.

To understand the significance of the hardware induced update rule, the behavior of Eq. 8 is described below for three different device switching characteristics, as illustrated in Fig 1. For the first device, that changes the conductance in a linear fashion and has the same value for the positive and negative branches, the hardware induced update rule simplifies back to the desired SGD update rule as $F(w) = 1$ and $G(w) = 0$. This is the case usually considered as the ideal device behavior required for good convergence. For the second device, that changes conductance in a nonlinear but symmetric fashion for both up and down branches, then again $G(w)$ term drops, and Eq. 8 simplifies to a form $w \leftarrow w - \eta \Delta w F(w)$, where more specifically $F(w) = 1 - 1.66w$ for the example device illustrated in Fig 1B. Although this update rule is different from the original SGD update rule, the existence of $F(w)$ only modifies the effective learning rate and therefore does not affect the convergence. Indeed, empirically it is shown that RPU devices need to have only symmetrical switching characteristics and the linearity is not required for good convergence [12] [20] [19] [26]. Only if updates are performed on two separate devices that change their conductances monotonically, such as PCM devices with one sided switching, then pair-wise matching and linearity are mandated to satisfy the symmetry requirement [27]. Finally, for the third device with asymmetric device switching characteristics the hardware induced update rule becomes $w \leftarrow w - \eta \Delta w - \eta |\Delta w| G(w)$. Since $|\Delta w|$ can only be non-negative, the last term is totally dictated by the functional form of $G(w)$ and appears due to an unintentional energy term introduced into optimization objective by the underlying hardware behavior. For the example device shown in Fig 1C where $G(w) = 1.66w$, the hardware induced update rule becomes $w \leftarrow w - \eta \Delta w - \eta |\Delta w|(1.66w)$ which corresponds to an optimization objection that is a combination of the original problem with an additional quadratic energy term $w^2$. This is similar to adding $\ell_2$ regularization term into the optimization objective but unfortunately its magnitude cannot be controlled and more strictly its amplitude is proportional to the updates $|\Delta w|$. This imposes a competition between the original optimization objective of the neural network and an internal energy term due to device characteristic; providing theoretical justification to the empirically observed poor training results obtained for non-symmetric RPU devices.

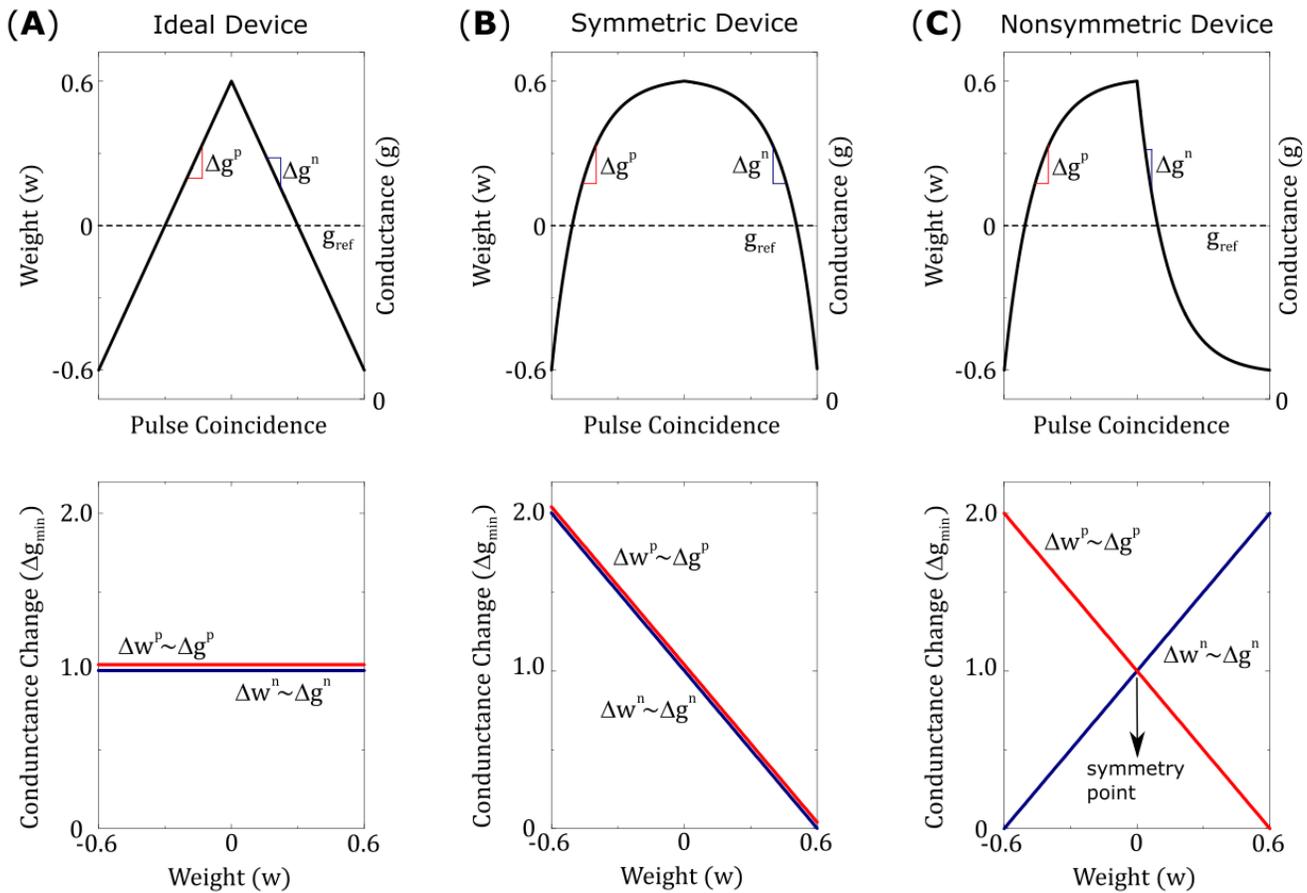

**Figure 1.** Three different device switching characteristics are illustrated. **(A)** Ideal device: Conductance increments and decrements are equal in size and do not depend on device conductance. **(B)** Symmetric device: Conductance increments and decrements are equal in strength, but both have a dependence on device conductance. **(C)** Nonsymmetric device: Conductance increments and decrements are not equal in strength and both have different dependencies on device conductance. However, there exists a single point that the strengths of the conductance increment and decrement are equal. This point is called the symmetry point and for the illustrated example matches the reference device conductance and hence happens at $w = 0$.

Note that even for the asymmetric device illustrated in Fig 1C there exists a single point that the strengths of the conductance increment and decrement are equal. This point is called the symmetry point of the updated device and it may correspond to any weight value (not necessary to zero as illustrated in Fig 1C) due to the device-to-device variations. As we show next there exists a method, so called *symmetry point shifting* technique, that can guarantee that the symmetry point of the updated device matches the conductance of the corresponding reference device and hence satisfying the condition $G_{ij}(w_{ij} = 0) \cong 0$ for all elements of the matrix. However, the behavior of $G(w)$ away from zero is still dictated by the updated device characteristics and for actual hardware implementations of RPU devices, each device would show different $G(w)$ characteristics due to device variability. Combination of device variability and conductance state dependent updates makes it impossible to compensate for this asymmetric behavior for individual devices without breaking the parallel nature of the array operations. However, the "Tiki-Taka" algorithm, as we describe below, eliminates the undesired effects of the device asymmetry for realistic RPU devices without breaking the array parallelism during training.

**Symmetry Point Shifting Technique**

The first step of the symmetry point shifting techniques is to apply a sequence of alternating (positive and negative) update pulses to the whole array all in parallel. In an alternating pulse sequence, the two consecutive pulses eliminate the $\eta \Delta w_{ij} F_{ij}(w_{ij})$ term from Eq. 8 and the dynamics of the whole array is dictated by the individual $G_{ij}(w_{ij})$ terms. The behavior of $G_{ij}(w_{ij})$ is expected to be different for each device due to the device variability and initial conductance variations, however, after sufficiently long sequence of pulses is applied, at steady state all elements are expected to converge to a point where $G_{ij}(w_{ij}) \cong 0$, although the corresponding weight value is not necessarily at zero, $w_{ij} \neq 0$. This behavior is expected for most physically plausible RPU devices. For instance, it is not realistic to expect alternative pulse sequence to give divergent conductance behavior, and instead, two consecutive pulses would push the conductance of the updated device towards the symmetry point $s_{ij}$ at which the up and down conductance changes are equal in strength and satisfy $\Delta g_{ij}^p(s_{ij}) = \Delta g_{ij}^n(s_{ij})$. As shown in Fig 1C, if the device conductance is smaller than the symmetry point ($g_{ij} < s_{ij}$) then the conductance increments are stronger than the decrements ($\Delta g_{ij}^p > \Delta g_{ij}^n$), and similarly, if the device conductance is larger than the symmetry point ($g_{ij} > s_{ij}$) then the conductance decrements are stronger than the increments ($\Delta g_{ij}^p > \Delta g_{ij}^n$). Therefore, independent of the initial conductance values, this alternating pulse sequence pushes all device conductances towards the corresponding symmetry points, as illustrated in Fig 2, eventually setting the conductance of each updated RPU device close to its symmetry point $s_{ij}$ for the whole array. As a second and last step these conductance values $s_{ij}$ are transferred to the corresponding reference devices so that $g_{ij,ref} \cong s_{ij}$ and hence $G_{ij}(w_{ij} = 0) \cong 0$ for all elements in the matrix. Since this is a onetime cost the conductance transfer can be performed iteratively in a closed loop fashion to overcome hardware limitations.

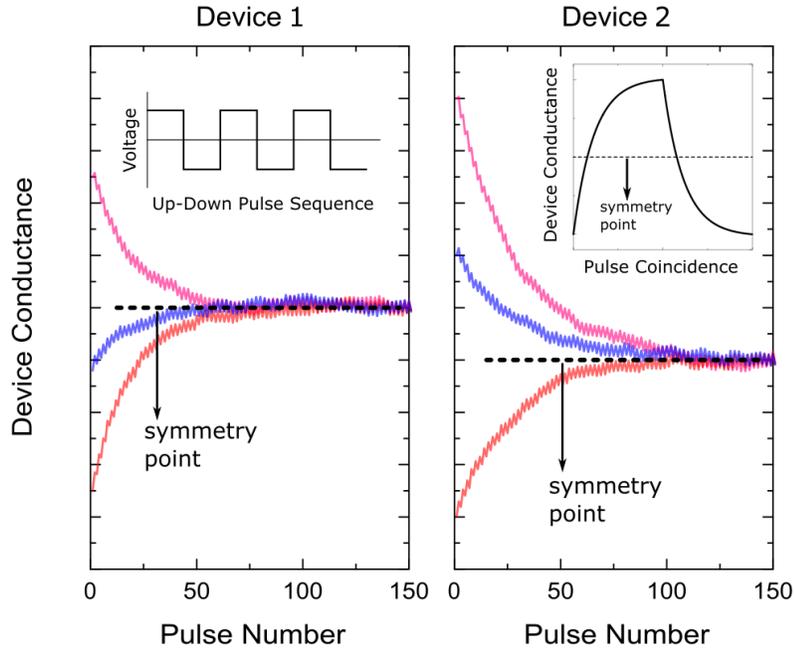

**Figure 2.** Illustration of the symmetry point shifting technique. Response of two separate devices to the alternating (up and down) pulse sequence starting from different initial conductance values.

## "Tiki-Taka" Algorithm

In "Tiki-Taka", each weight matrix of the neural network is represented by a linear combination of two matrices

$$W = \gamma A + C \tag{9}$$

where $A$ is the first matrix, $C$ is the second matrix and $\gamma$ is a scalar factor that controls the mixing of the two matrices. The elements of $A$ and $C$ matrices, $a_{ij}$ and $c_{ij}$ respectively, are also encoded by a pair of devices and we use upper left superscripts $a$ and $c$ consistently to refer to the properties of the elements (and devices) in $A$ and $C$. For instance, $^{a}g_{ij}$ and $^{c}g_{ij}$ denote the conductance values stored on devices used for updates, and similarly $^{a}g_{ij,ref}$ and $^{c}g_{ij,ref}$ denote the conductance values stored on devices used as references corresponding to $i^{th}$ column and $j^{th}$ row. For "Tiki-Taka" to be successful, important criteria, $^{a}G_{ij}(a_{ij} = 0) \cong 0$ for all elements of $A$, must be realized by the hardware. Therefore, we assume that the symmetry point shifting technique is applied to $A$ before starting the training procedure described below.

## Training Procedure

To simplify the exposition of the key idea, we shall omit for now the non-linear activation functions. In its most general form, the weight matrix $W$ is a linear combination of two matrices $A$ and $C$ and $\gamma$ is a scalar factor that controls the mixing of the two matrices. During training, the weight updates are accumulated on $A$ that has symmetric behavior around the zero point, and then moved to $C$. The operations performed during "Tiki-Taka" is summarized in Table 2 along with the ones performed during the SGD\BP algorithm for comparison.

**Table 2. Operations for SGD\BP algorithm and "Tiki-Taka" algorithm on a single layer**

| SGD\BP Algorithm | "Tiki-Taka" Algorithm |
|---|---|
| for each data in training dataset<br>{<br><br>(1)   $y = Wx$<br>(2)   $z = W^\top \delta$<br>(3)   $W \leftarrow W + \eta\,(x \otimes \delta)$<br>} | k = 0<br>for each data in training dataset<br>{<br>   k = k + 1<br>(1)   $y = (\gamma A + C)x$<br>(2)   $z = (\gamma A + C)^\top \delta$<br>(3)   $A \leftarrow A + \eta\,(x \otimes \delta)$<br>  if (k == ns)<br>  {<br>    k = 0<br>    t = t + 1<br>(4)   $v = A u_t$<br>(5)   $C \leftarrow C + \lambda\,(u_t \otimes v)$<br>  }<br>} |

For simplicity, only the operations performed on the weight matrices are shown.

The conventional SGD\BP algorithm is composed of three cycles: (1) forward, (2) backward and (3) weight update; whereas for "Tiki-Taka" there exist five cycles: (1) forward, (2) backward, (3) update $A$, (4) forward $A$, and (5) update $C$. The first two (forward and backward) cycles of the "Tiki-Taka" algorithm are identical to the ones in SGD\BP, as "Tiki-Taka" also uses the conventional BP algorithm to calculate the gradients. However, instead of using a single weight matrix, the linear combination of two matrices is used to perform the forward and backward computations. The third (update $A$) cycle is same as the weight update cycle of the SGD\BP algorithm and the update operation on $A$ is performed using the outer product of the two vectors that are used in the forward and the backward cycles. These three cycles are repeated $ns$ times before the fourth and fifth cycles of the "Tiki-Taka" algorithm are performed. In the fourth (forward $A$) cycle a vector-matrix multiplication is performed on $A$ using an input vector $u_t$. We discuss different choices for $u_t$ later but in its most simple form $u_t$ is a single column of an identity matrix (a one-hot encoded vector) where for each artificial time step a new column is used in a cyclic fashion and the sub index $t$ denotes that time index. This operation effectively reads a single column of $A$ into the resulting vector $v$. In the fifth (update $C$) cycle, $C$ is updated using the outer product of the same two vectors $u_t$ and $v$ from the fourth cycle. This update operation changes only the elements corresponding to a single column of $C$ proportional to the values stored in $A$, and $\lambda$ is the learning rate used for updating $C$. Note that since different $u_t$ is used at different time steps eventually all elements of $C$ get updated.

In this algorithm the updates performed on $A$ accumulate the gradients from different data examples and therefore $A$ is actively used. In contrast, the updates on $C$ are very sparse and only a single column of $C$ is updated while the remaining elements are kept constant. Therefore, there is a big difference in the update frequency of the elements in these two matrices and $C$ learns on a much slower time scale only using the information accumulated on $A$. As described above, the gradient accumulation happening on $A$ have artifacts due to the hardware induced update rule, however, thanks to the symmetry point shifting technique the sign of the average gradient information is very likely to be correct (up to a limit that is dictated by the hardware noise). For instance, any kind of randomness in updates due to the random sampling of the data examples pushes the elements of $A$ towards zero while the true average gradients push them away from zero. Therefore, when the elements of $A$ are read, all elements are likely to have the correct sign information of the accumulated gradients although the amplitudes are probably underestimated. This information is then transferred to $C$ which effectively grows the total weight in the correct direction that minimizes the objective function. At the end of the training process, at the steady state, (independent of the choice of $\gamma$ value) we expect the elements of $C$ would get very close to a point in space, $c_{ij} \cong w_{ij,opt}$, where the original objective function is in its local minima and the elements of $A$ would be close to zero, $a_{ij} \cong 0$. This is indeed a stable point for the coupled system to the first order. When $c_{ij} \cong w_{ij,opt}$, by definition the average gradients from different data samples are close to zero, $\langle \Delta w_{ij} \rangle \cong 0$, but since $\langle |\Delta w_{ij}| \rangle$ is always finite due to stochastic data sampling the hardware induced update rule for $A$ drives all elements towards zero, $a_{ij} \cong 0$, which in return diminishes the updates on $C$. We note that the hardware induced update rule for $C$ also has artifacts that repels $c_{ij}$ away from $w_{ij,opt}$, however, these updates are sparse and happens across much slower time scales, rendering such artifacts negligible.

In contrast, for the SGD\BP algorithm even if the weights somehow get close to an optimum point corresponding to a local minimum, $w_{ij} \cong w_{ij,opt}$, the randomness in the updates pushes the weights away from the optimum point towards the symmetry point (or towards zero if symmetry point shifting technique is applied). Using the same arguments presented above, at a local minimum not only the average gradients from different data samples are close to zero, $\langle \Delta w_{ij} \rangle \cong 0$, but also it is guaranteed that $\langle |\Delta w_{ij}| \rangle > 0$ due to random data sampling. Therefore, at an optimum point the hardware induced updates are totally predominated by the nonsymmetric device switching characteristics, $G_{ij}(w_{ij})$, and therefore optimum points are not stable points for the SGD\BP algorithm running on RPU hardware. The only stable points that SGD\BP can find are the ones that has a tension between original optimization objective and the internal energy term due to device asymmetry, such that $\langle \Delta w_{ij} \rangle F_{ij}(w_{ij}) \cong -\langle |\Delta w_{ij}| \rangle G_{ij}(w_{ij})$, and therefore give non satisfactory training results. The artifacts of the hardware induced update rule for the SGD\BP algorithm are mitigated using the "Tiki-Taka" algorithm where the optimum points of the original objective function are turned into stable points for the coupled dynamical system. Therefore, this new training approach is expected to give superior results compared to the SGD\BP algorithm when running on RPU hardware.

**RESULTS**

To test the validity of the proposed "Tiki-Taka" algorithm we performed DNN training simulations on three different network architectures: (1) **FCN-MNIST** -- a fully connected network trained on MNIST dataset, (2) **CNN-MNIST** -- LeNet5 like convolutional neural network trained on MNIST dataset and (3)

**LSTM-WP** -- a doubly stacked LSTM network trained on Leo Tolstoy's War and Peace (WP) novel. For all these three networks, the training performance of the SGD\BP algorithm with realistic RPU device specifications was studied carefully in previous publications [12] [19] [26]. It was shown that a very tight symmetry requirement is needed to achieve training accuracies comparable to the ones achieved with high precision floating point numbers. Here, we use the same network settings from those publications, such as the activations, the layer sizes and the layer mappings onto the arrays; and follow a similar methodology for the RPU models, such that they capture the variations and non-idealities of the RPU devices as well as the peripherical circuitry driving the arrays. However, we emphasize that different from those studies, here we use a significantly nonsymmetric device switching behavior as described below to evaluate the performance of the "Tiki-Taka" algorithm.

**RPU Baseline Model**

The RPU-baseline model uses the stochastic update scheme in which the numbers that are encoded from the periphery ($x_i$ and $\delta_j$) are implemented as stochastic bit streams. Each RPU device then performs a stochastic multiplication via simple coincidence detection. In our simulation tool, each coincidence event triggers an increment or decrement in the corresponding device conductance using a device switching characteristics. As a baseline model we use a nonsymmetric device behavior, similar to one shown in Fig 1C, and this behavior introduces a weight update (conductance change) that depends on the current weight value (current device conductance) and the direction of the update. The dependence of the incremental weight updates for both branches are assumed to be linear: for the positive branch $\Delta w_{min}^p(w) = \Delta w_{min0}(1 - slope^p \times w)$ and for the negative branch $\Delta w_{min}^n(w) = \Delta w_{min0}(1 + slope^n \times w)$, where $slope^p$ and $slope^n$ are the slopes that control the dependence of the weight changes on the current weight values, and $\Delta w_{min0}$ is the weight change due to a single coincidence event at the symmetry point corresponding to the zero weight value. For the device illustrated in Fig 1C, $\Delta w_{min0} = 0.001$, $slope^p = 1.66$ and $slope^n = 1.66$. For the baseline model these values are used only as averages and for each RPU device there exists three unique parameters that are sampled from Gaussian distributions at the beginning of the training and then used throughout the training. To capture the device-to-device variations all Gaussian distributions have significant standard deviations, and relative to the mean values mentioned above 30%, 25% and 25% standard deviations are used for parameters $\Delta w_{min0}$, $slope^p$ and $slope^n$ respectively. Moreover, to capture the cycle-to-cycle variations for each coincidence event an additional 30% Gaussian noise is introduced to $\Delta w_{min}^p$ or $\Delta w_{min}^n$ relative to their expected values before incrementing or decrementing the corresponding weight value. In this model, the weight saturations, corresponding to conductance saturations, are automatically taken into account due to the weight dependent updates, and the weight values cannot grow bigger than $1/slope^p$ or smaller than $-1/slope^n$. For a nominal device this corresponds to weight bounds between $-0.6$ and $0.6$ but note that for each device $slope^p$ and $slope^n$ are sampled independently and therefore they don't necessary match. In the context of "Tiki-Taka", since we use a new set of random variables for each device model there is no correlation between the elements of $\boldsymbol{A}$ and $\boldsymbol{C}$, and in this context the weight changes refer to the changes that occur in the elements of $\boldsymbol{A}$ and $\boldsymbol{C}$. Note that because of the dependencies assumed in the model, the expected weight changes for the positive and negative branches are equal in strength at zero weight value, $\Delta w_{min}^n(w=0) = \Delta w_{min}^p(w=0) = \Delta w_{min0}$. Therefore, this model assumes that the symmetry point shifting technique is applied perfectly both to $\boldsymbol{A}$ and $\boldsymbol{C}$, such that all reference device conductances are set to the symmetry point of the corresponding device used for updates. Later we relax this assumption to test the tolerance of "Tiki-Taka" to the symmetry point variations.

We emphasize that the chosen mean value for $slope^p = slope^n = 1.66$ that control the device asymmetry is the largest possible value that can be used without introducing any side effects. For instance, it is shown in Ref [12] that the acceptable criteria for the weight bounds is between $-0.6$ and $0.6$ and this range is consistently used in Refs [19] [26]. Therefore, increasing the slope parameters beyond 1.66 would limit the weights into a range that is tighter than the acceptable criteria. Although "Tiki-Taka" is expected to deal with the device asymmetry, it cannot improve over these weight bounds resulting in side effects into the training. The chosen mean value for $slope^p = slope^n = 1.66$ is therefore the most aggressive asymmetric device switching behavior that can used without violating the other RPU specs derived in Ref [12].

In addition to the non-idealities mentioned above, for any real hardware implementations of RPU arrays the results of the vector matrix multiplications will be noisy as well and this noise is considered by introducing an additive Gaussian noise, with zero mean and standard deviation of $\sigma = 0.06$. Moreover, the results of the vector-matrix multiplications are bounded to a value of $|\alpha| = 12$ to account for signal saturation. The input signals are assumed to be between [-1, 1] with a 7-bit input resolution, whereas the outputs are quantized assuming a 9-bit ADC. Although the input signals going into the array and the output signals coming from the arrays are bounded, we use noise management and bound management techniques described in Ref [19]. We note that apart from the nonsymmetric update behavior used for RPU devices, all other hardware constraints, such as variations, noise, limited resolutions and signal bounds, are identical to the ones used in publications [12] [19] [26].

**Fully Connected Network on MNIST (FCN-MNIST)**

The same network from Ref [12], composed of fully connected layers with 784, 256, 128 and 10 neurons, is trained with the standard MNIST dataset composed of 60,000 training images. For hidden and output layers $sigmoid$ and $softmax$ activations are used, respectively. For the floating point (FP) model, training is performed with the SGD algorithm using a mini-batch size of unity and a fixed learning rate of $\eta = 0.01$. As shown by open symbols in Figure 3, the FP-model reaches a classification error of 2.0% on the test data after 50 epochs of training. The same SGD based training using the RPU-baseline model however results in about 15% test error that is significantly higher than the error achieved by the FP-model. This is indeed expected, as the device characteristics in the RPU-baseline model is highly asymmetric and well above the acceptable device symmetry criteria described in Ref [12]. When the training is performed using "Tiki-Taka" for the same the RPU-baseline model, the test error drops back to a value close to 2%. This level of error is indistinguishable from the one achieved by the FP-model, and shows that in contrast to SGD, "Tiki-Taka" gives good training results even with highly nonsymmetric devices. We emphasize that the "Tiki-Taka" algorithm is no more sensitive to other hardware issues (such as stochastic updates, limited number of steps, noise, ADC, and DAC) than the SGD algorithm as the RPU-model captures all those hardware constraints.

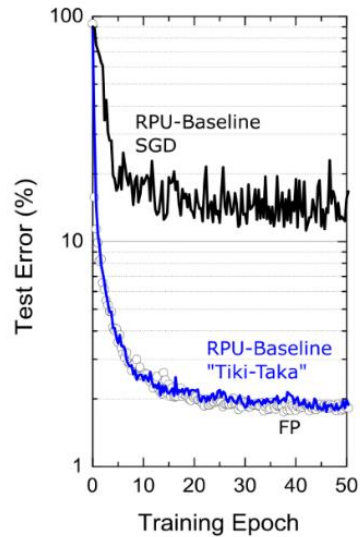

**Figure 3.** Test error of FCN on MNIST dataset. Open white circles correspond to a model where the training is performed using high precision floating point (FP) numbers and the SGD algorithm. Black and blue lines correspond to the RPU-baseline model where trainings are performed using the SGD algorithm and the "Tiki-Taka" algorithm, respectively.

Note that different from a single learning rate used for SGD, there exist additional hyperparameters for the "Tiki-Taka" algorithm, as illustrated in Table 2, namely the learning rates $\eta$ for updating $A$ and $\lambda$ for updating $C$, the parameter $\gamma$ controlling the mixing between $A$ and $C$, the parameter $ns$ controlling the period of updates performed on $C$, and the choice of $u_t$ used for the forward $A$ cycle. In order to present the best results possible by "Tiki-Taka" we performed training simulations at different hyperparameter settings. For the results presented in Figure 3 the learning rates $\eta$ and $\lambda$ are fixed at 0.01 and 0.02 respectively, the parameters $\gamma$ and $ns$ are set to unity, and for $u_t$ a fixed set of one-hot encoded vectors are used in a cyclic fashion. Additional training results studying the sensitivity of "Tiki-Taka" to some of these hyperparameters are shown in Fig 4 for the FCN-MNIST problem.

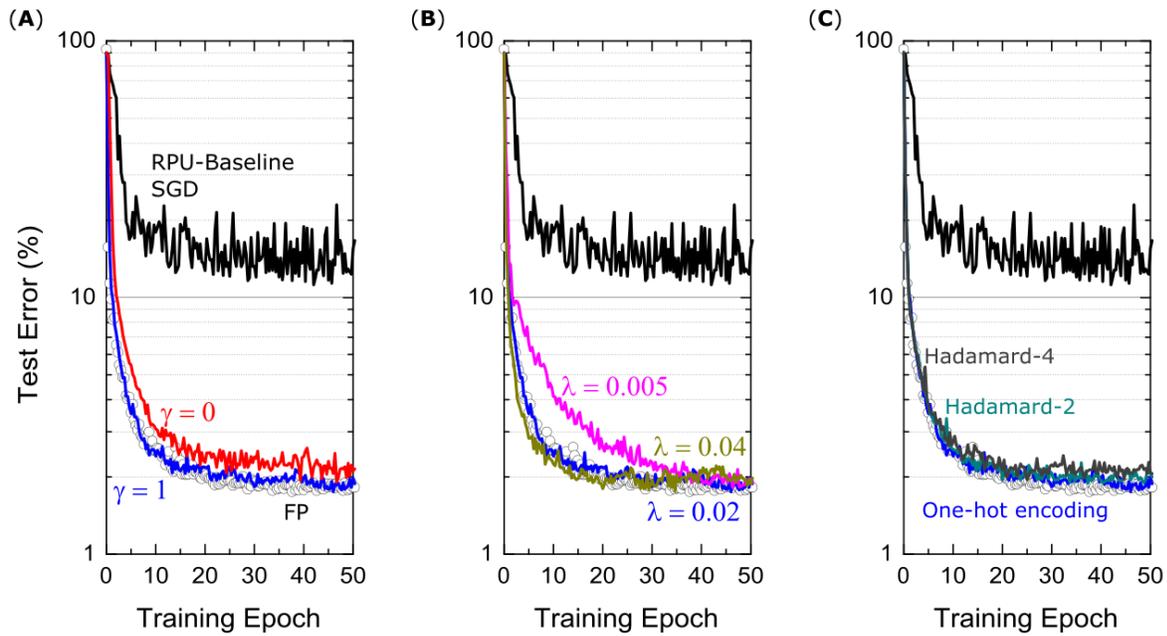

**Figure 4.** Test error of FCN on MNIST dataset trained using Tiki-Taka algorithm at different hyperparameter settings. **(A)** The comparison of the training results at two different mixing terms, $\gamma = 1$ and $\gamma = 0$, corresponding to blue and red curves respectively. **(B)** The comparison of the training results at three different learning rates on $C$ matrix, $\lambda = 0.005, 0.02$ and $0.04$, corresponding to magenta, blue and green curves respectively. **(C)** The comparison of training results at three different choices of $u_t$ vectors. Blue curves uses one-hot encoded vectors in a cyclic fashion. Cyan and gray curve respectively uses the vectors of Hadamard-2 and Hadamard-4 matrices in a cyclic fashion. For all figures, open white circles correspond to a model where the training is performed using high precision floating point numbers using the SGD algorithm, and black curve corresponds to the RPU-baseline model where the training is performed using the SGD algorithm.

In Fig 4A, the training results for cases $\gamma = 1$ and $\gamma = 0$ are shown while all other hyperparameters are unchanged. When $\gamma = 1$, the gradient updates happening on $A$ are directly visible in the next cycle while calculating the gradients corresponding to another image. In contrast, for $\gamma = 0$, the learning can happen only indirectly thorough the updates performed on $C$ that are sparse and less frequent. As described above we don't expect the steady state solutions to be any different between cases $\gamma = 1$ and $\gamma = 0$. Consistently, these two simulations show very similar training curves that improve significantly over the SGD training and reach test accuracies comparable to one achieved by the FP-model. However, the training process is governed by the dynamics of the coupled system and not by the equilibrium properties. Therefore, one may argue that the case $\gamma = 0$ learns slightly slower than case $\gamma = 1$ due to infrequent updates to explain the slight difference observed between the two cases for FCN-MNIST problem. Furthermore, these simulations also consider other possible hardware issues due to stochastic pulsing, variations as well as noise. Therefore, one may also expect case $\gamma = 0$ to perform better gradient estimations on $A$ before transferring that information on to $C$ and hence to show better training performance overall than case $\gamma = 1$. Although, there exist these interesting tradeoffs while choosing the mixing term, large improvements over the SGD training are consistently observed as illustrated in Fig 4A independent of $\gamma$.

In Fig 4B we present the training results at various learning rates $\lambda$ used to update $C$. It is clear that choosing a too large or a too small $\lambda$ values are both undesirable. In the case of a too small $\lambda$, where $\lambda \to 0$, "Tiki-Taka" implements the original SGD algorithm (assuming $\gamma = 1$). In this setting only $A$ learns using the same SGD algorithm on a weight space that is shifted by values stored in $C$, but, $A$ cannot not find good solutions because of the artifacts introduced by the hardware induced update rule. On the other extreme choosing a large $\lambda$ results in an unstable behavior for the coupled system. In our simulations, we try a few $\lambda$ values that are close to the learning rate $\eta$. We believe choosing similar learning rates keeps the updates on both systems comparable in strength and therefore the couple system can minimize the both objective functions simultaneously in a self-consistent fashion. The simulation results at three different $\lambda$ values, 0.005, 0.02 and 0.04, are show in Fig 4B, all of which are achieving comparable test errors at the end of 50 epochs.

Other important hyperparameters of the "Tiki-Taka" algorithm are the $u_t$ vectors used while doing a forward pass on $A$ and $ns$, the period used to update $C$. Note that there exist three weight matrices for FCN-MNIST, each having the dimensions of 256x785, 128x257 and 10x129 including the bias terms, where each weight matrix is represented by a pair of matrices $A$ and $C$ in the "Tiki-Taka" algorithm. Therefore, even when $ns = 1$ and a fixed set of one-hot encoded vectors is used for $u_t$, for the first layer it takes 785 images for all elements of $C$ to get updated only once. Similarly, 257 and 129 images are required for the following layers. Larger $ns$ values can be used to reduce the number of updates happening on $C$ compared to $A$ and for each layer $ns$ can be chosen independently. Increasing the update period on $C$ makes the artifacts of the hardware induced update even less important for $C$. However, note that the randomness of the updates on $A$ tends to push the values of $A$ towards zero due to the hardware induced update rule and hence slowly erases true gradient information accumulated on $A$ from earlier time steps (images). Therefore, increasing $ns$ beyond a certain value would not make the gradient accumulation more accurate and therefore there exists an upper bound on how large $ns$ can be increased meaningfully. On the contrary, one may want to perform updates more often than the case supported by $ns = 1$ in order to use the hardware induced updates for regularization purposes. As illustrated before, the randomness in the updates attracts the corresponding matrices towards zero which has effects similar to the $\ell_2$ regularization for some specific device switching characteristics, but the strength wasn't controllable for the SGD algorithm. In contrast, in "Tiki-Taka" we can control the strength of this term by performing updates on $C$ using various $u_t$ vectors. For instance, instead of using a set of one-hot encoded vectors in a cyclic fashion, the vectors of the normalized Hadamard-2 matrix padded with zeros, such as $\left[\frac{1}{2} \ \frac{1}{2} \ 0 \ 0 \ ...\right]$ $\left[\frac{1}{2} \ -\frac{1}{2} \ 0 \ 0 \ ...\right], \left[0 \ 0 \ \frac{1}{2} \ \frac{1}{2} \ ...\right]$ and so on, can be used in a cyclic fashion. This results in twice more updates on each element of $C$, yet a similar information is transferred from $A$ to $C$. Because of the cancellations happening between two back-to-back updates on $C$, it would experience a stronger regularization towards zero thanks to the hardware induced update rule. Examples of the training performed using the vectors of the Hadamard-2 and Hadamard-4 matrices are shown in Fig 4C. These examples show that similar information can be transferred from $A$ to $C$ independent of the choice of $u_t$ and yet the choice of $u_t$ can be used as a knob to control the regularization term. We note that FCN-MNIST is a simple problem and does not overfit and hence does not require regularization, however, it is important to understand the consequences of different hyperparameters, so they can be tuned properly when they are really needed for large scale networks.

**Convolutional Neural Network on MNIST (CNN-MNIST)**

The CNN network used here is the same network from Ref [19] and is composed of two convolutional and two fully connected layers. The first two convolutional layers use $5 \times 5$ kernels each having 16 and 32 kernels, respectively. Each convolutional layer is followed by a subsampling layer that implements the max pooling function over non-overlapping pooling windows of size $2 \times 2$. The output of the second pooling layer, consisting of 512 neuron activations, feeds into a fully connected layer consisting of 128 $tanh$ neurons, which is then connected into a 10-way $softmax$ output layer. Including the biases there exist four weight matrices with dimensions of $16 \times 26$ and $32 \times 401$ for the first two convolutional layers and, $128 \times 513$ and $10 \times 129$ for the following two fully connected layers.

We note that different from the fully-connected layers, convolutional layers have weight sharing that changes the vector operations performed on the weight matrices to matrix operations that are implemented as a series of vector operations on the arrays as described in Ref [19]. For "Tiki-Taka" this means that the first 3 cycles corresponding to the convolutional layers are now matrix operations and can be written as $\boldsymbol{y} = (\gamma \boldsymbol{A} + \boldsymbol{C})\boldsymbol{X}$, $\boldsymbol{z} = (\gamma \boldsymbol{A} + \boldsymbol{C})^T \boldsymbol{\Delta}$, and $\boldsymbol{A} \leftarrow \boldsymbol{A} + \eta\,(\boldsymbol{X} \otimes \boldsymbol{D})$, where $\boldsymbol{X}$ and $\boldsymbol{\Delta}$ are the inputs and the errors feed into the weight matrices in the forward and backward directions, while the 4$^{\text{th}}$ and 5$^{\text{th}}$ cycles remains as before. The weight sharing factors for the two convolutional layers are 576 and 64, respectively. Therefore, when $ns = 1$ and a one-hot encoded vector is used as $\boldsymbol{u}_t$, the $\boldsymbol{A}$ matrix of the first convolutional layer is updated 576 times before a single column of $\boldsymbol{C}$ is updated. Similarly, for the second convolutional layer $\boldsymbol{A}$ is updated 64 times, before $\boldsymbol{C}$ gets a single column update. All other operations remains identical for fully connected layers.

CNN-MNIST simulation results are shown in Fig 5. For the FP-model, trained with $\eta = 0.01$ and mini-batch of unity, the network achieves a test error of 0.8%. However, when RPU-baseline model is trained with the SGD algorithm, the test error is very high at around 8%. This large discrepancy from the FP-model significantly drops when the RPU-baseline model is trained with the "Tiki-Taka" algorithm, resulting in 1.0% test error. To understand the cause of this remaining 0.2% offset from the FP-model, we repeat the SGD training assuming a device model with perfect symmetry ($slope^p = slope^n = 0$ for all devices) but with the remaining hardware constraints. This perfectly symmetric case trained with the SGD algorithm gives a test accuracy not any better than the one achieved by the nonsymmetric case trained with "Tiki-Taka", suggesting that the remaining 0.2% discrepancy from the FP-model is due to other hardware constraints and not due to the device asymmetry. These results further highlight the power of this new training technique that compensated for device asymmetry at the algorithm level.

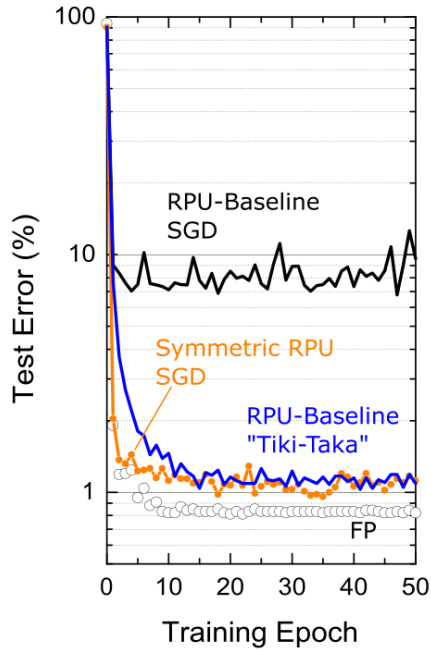

**Figure 5.** Test error of CNN on MNIST dataset. Open white circles correspond to a model where the training is performed using high precision floating point (FP) numbers using the SGD algorithm. Black and blue lines correspond to the RPU-baseline model where trainings are performed using the SGD algorithm and the "Tiki-Taka" algorithm, respectively. The orange points\line correspond to SGD based training of a RPU model where all devices are perfectly symmetric while all other variations are identical to RPU-baseline model.

**Sensitivity to Symmetry Point Variations**

The simulations results presented so far assume the symmetry point shifting techniques is applied perfectly and hence the update strength for positive and negative branches are equal in strength at $w = 0$. It is clear that the symmetry point shifting technique cannot be perfect due to hardware limitations: To test the tolerance of the "Tiki-Taka" algorithm to the symmetry point variations, we performed training simulations by relaxing this assumption such that the condition $\Delta w_{min}^n(w_s) = \Delta w_{min}^p(w_s)$ happens at a weight value $w_s$ that is different for each element in $\boldsymbol{A}$ and $\boldsymbol{C}$. This is simply achieved by setting the reference device conductance different from the symmetry point of the corresponding device used for the updates in the RPU baseline model. The simulation results for both FCN-MNIST and CNN-MNIST are presented in Fig 6, where $w_s$ value for each device is sampled from a Gaussian distribution with zero mean but varying standard deviation, $\sigma_{ws}$. When the standard deviation of the distribution is $\sigma_{ws} \leq 0.01$, the training results are indistinguishable from the one achieved with no variations, and therefore, these results provide $\sigma_{ws} = 0.01$ as the acceptable threshold value for the symmetry point variations.

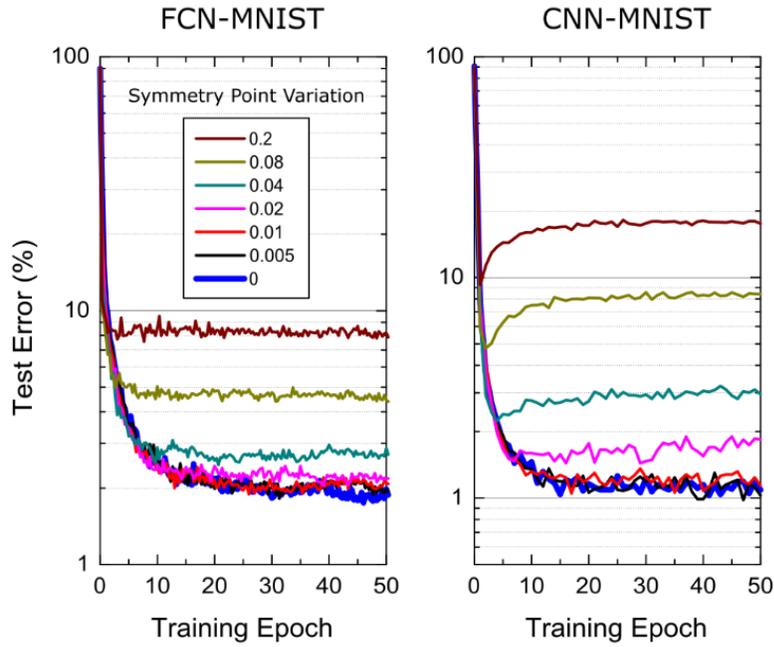

**Figure 6.** Sensitivity of the "Tiki-Taka" algorithm to the symmetry point variations for FCN-MNIST and CNN-MNIST. The same hyperparameter settings from Fig 1 and Fig 5. are used for FCN-MNIST and CNN-MNIST, respectively. Different colors correspond to RPU-baseline models at different symmetry point variations.

It is important that this acceptable threshold value of $\sigma_{ws} = 0.01$ is achieved by the symmetry point shifting technique. The symmetry point shifting technique may introduce two sources of noise while matching the reference device conductance to the symmetry point of the updated device: (1) the noise in the converge to the symmetry point of the updated device and (2) the noise in the conductance transfer to the reference device. Given than the weight change due to a single coincidence event at the symmetry point is about $\Delta w_{min0} = 0.001$, which is 10 times smaller than the threshold $\sigma_{ws} = 0.01$, the alternating pulse sequence would result in convergence to the symmetry point that is much smaller than this acceptable threshold. Furthermore, using the ratio of this acceptable threshold value $\sigma_{ws} = 0.01$ to a nominal weight range of 1.2, this specification can be mapped to physical quantities. For instance, the matching of the reference device conductance to the symmetry point of the updated device must be accurate within a few percent compared to the whole conductance range. Therefore, after this initial converge, the stored conductance on the update device needs to be copied to the reference device within a few percent accuracy. Given that this is a onetime cost, this conductance transfer can be performed using a closed loop programming while achieving this required few perfect matching. Therefore, we emphasize that the acceptable threshold for symmetry point variations can be achieved with the symmetry point shifting technique and does not introduced any additional constraints to the required device specifications.

**LSTM Network on War and Peace Dataset (LSTM-WP)**

As a third example, the validity of the "Tiki-Taka" algorithm is tested on a more challenging LSTM network. This network is composed of 2 stacked LSTM blocks each with a hidden state size of 64 and it is identical to one studied in Ref [26]. As a dataset Leo Tolstoy's War and Peace (WP) novel is used and it is split into training and test sets as 2,933,246 and 325,000 characters with a total vocabulary of 87 characters. Following the same mapping described in Ref [26] results in 3 different weight matrices with sizes $256 \times (64 + 87 + 1)$ and $256 \times (64 + 64 + 1)$ for the two LSTM blocks and a third matrix of size $87 \times (64 + 1)$ for the last fully connect layer before the $softmax$ activation. Note that each matrix is now mapped to 2 separate matrices in "Tiki-Taka". The simulation results corresponding to the SGD algorithm and "Tiki-Taka" for various RPU models are shown in Fig 7. For all models the training is performed using fixed learning rate $\eta = 0.005$, mini-batch of unity and time unrolling steps of 100. Additionally, the hyperparameters of the "Tiki-Taka" algorithm are $\lambda = 0.005$, $\gamma = 1$, $ns = 5$, and for $\boldsymbol{u_t}$ one-hot encoded vectors are used in a cyclic fashion.

The simulation results presented in Fig 7A for LSTM-WP are qualitatively in good agreement with the ones presented for FCN-MNIST and CNN-MNIST above. For instance, the RPU-baseline model trained with the SGD algorithm results in a test error (cross-entropy loss) significantly larger than the one achieved by the FP-model. However, the same RPU-baseline model performs much better when "Tiki-Taka" is used for training, further validating this new approach for training DNNs. The perfectly symmetric case trained with the SGD algorithm is also shown as a comparison, and interestingly, it shows quantitative differences compared to ones presented for FCN-MNIST and CNN-MNIST: First, the perfectly symmetric case trained with the SGD algorithm cannot reach the level of accuracy achieved by the FP-model. Second, the RPU-baseline model trained with "Tiki-Taka" cannot reach the level of accuracy achieved by the perfectly symmetric case trained with the SGD algorithm. The former is understandable as it is shown in Ref [26] that LSTM networks are more challenging to train on crossbar arrays; and even for perfectly symmetric devices, FP model accuracies cannot be reached due to the limited number of states on RPU devices. Given that "Tiki-Taka" only addresses issues arising due to device asymmetry, it is not expected to reach the same level of accuracy of the FP model. It is only reasonable to expect it to perform at the same level of accuracy achieved by the perfectly symmetric case trained with the SGD algorithm, as all other hardware constraints are the same. Therefore, it is worth investigating the quantitative difference observed between the RPU-baseline model trained with the "Tiki-Taka" algorithm and the perfectly symmetric case trained with the SGD algorithm.

When the same RPU models, with significant device asymmetry, are used, it is clear that the training results using "Tiki-Taka" outperforms the results achieved by the SGD algorithm. This relaxes the acceptable device symmetry requirement by a large margin at equivalent accuracy, however, it is also obvious that reducing device asymmetry improves the training accuracy of the "Tiki-Taka" algorithm if the accuracy is already far from ideal to start with. Therefore, one can easily blame the very aggressive device asymmetry used in the RPU-baseline model to explain the quantitative difference observed between the "Tiki-Taka" algorithm and the perfectly symmetric case trained with the SGD algorithm. Trivially this deficit can be minimized by using a less asymmetric device switching characteristics (data not shown). However, there exist other hardware issues that may hinder the convergence, and more interestingly there may exists different tradeoffs between device switching characteristics and other hardware limitations for "Tiki-Taka" that are otherwise not applicable to the SGD algorithm.

To understand whether other existing hardware limitations play a role in the convergence of the "Tiki-Taka" algorithm we performed additional simulations using the same device model but assuming different hardware settings at the peripheral circuits. For instance, the simulation results presented in Fig 7B assume that the noise level for the vector-matrix multiplications is reduced by 10x from its original value in the RPU-baseline model. This reduction does not affect the performance of SGD based training both for the RPU-baseline model and the perfectly symmetric case. However, "Tiki-Taka" based training improves and the difference observed in Fig 7A between the RPU-baseline model trained with "Tiki-Taka" and the perfectly symmetric case trained with the SGD algorithm disappears in Fig 7B. More interestingly, in Fig 7C when we repeat the same experiment using an RPU-baseline model where only the noise spec of the forward $A$ cycle in the "Tiki-Taka" algorithm is reduced by 10x, the training result remains unchanged and are very close to the perfectly symmetric case trained with the SGD algorithm. These simulation results show that the noise introduced during the transfer of the information accumulated on $A$ to $C$ may play a role in the convergence of the "Tiki-Taka" algorithm.

We emphasize that the hardware induced update rule for $C$ also has artifacts that push the elements of $C$ away from the optimum points at equilibrium. Although these artifacts are less important compared to the SGD algorithm, increased noise in updating $C$ due to the randomness in reading $A$ clearly impacted the training accuracy of the "Tiki-Taka" algorithm as illustrated above. Therefore, in order to further filter the updates happening on $C$, we changed the update $C$ cycle of the "Tiki-Taka" algorithm as follows $C \leftarrow C + \lambda \, (u_t \otimes f(v))$, where $f(v)$ is a pointwise thresholding function that returns $v$ only if $|v| > T_v$ and otherwise zero. The simulation result of this modified "Tiki-Taka" algorithm for the RPU-baseline model is shown in Fig 7C, where $T_v$ is set to the one-sigma value of the read noise from the baseline model, $T_v = 0.06$. Although the improvement is not large, this filtering approach performs slightly better than the original unfiltered version and it suggests that there may exists other strategies that may outperform this simple thresholding-based filtering.

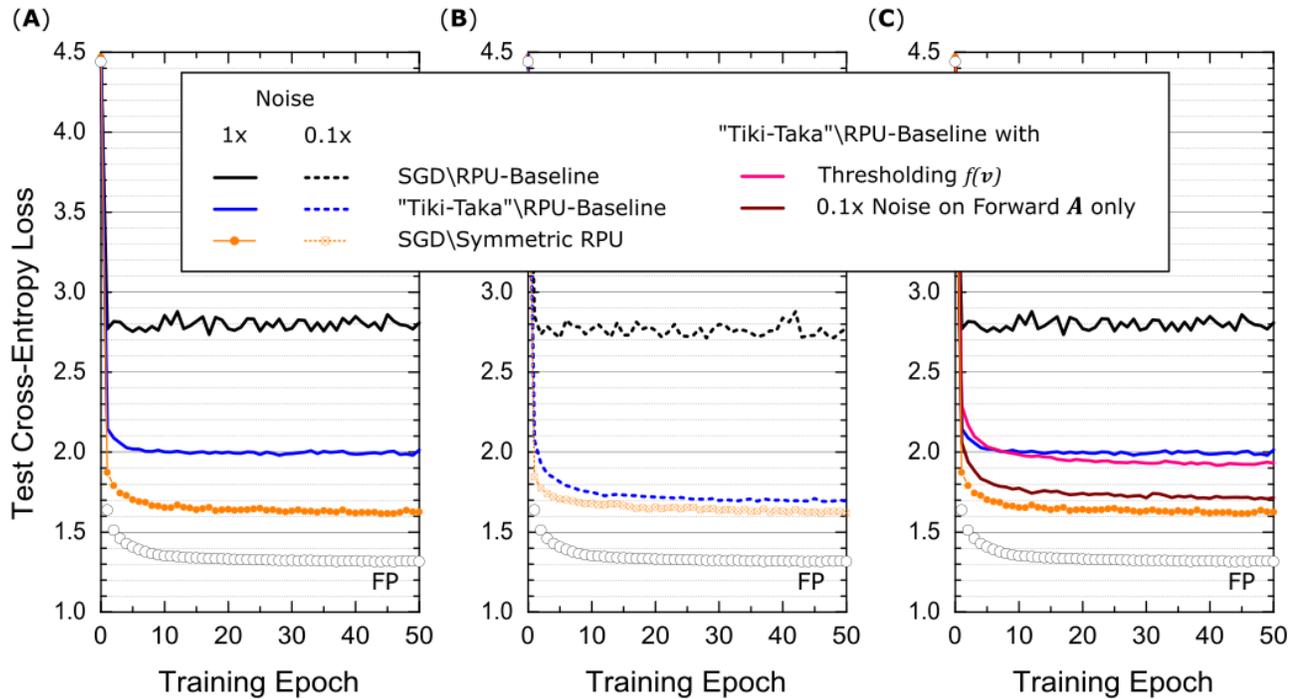

**Figure 7.** Test cross-entropy error for LSTM network trained on WP dataset. Open white circles correspond to a model where the training is performed using high precision floating point (FP) numbers using the SGD algorithm. **(A)** Black and blue lines correspond to the RPU-baseline model where trainings are performed using the SGD algorithm and the "Tiki-Taka" algorithm, respectively. The orange points\line correspond to SGD based training of an RPU model where all devices have a perfectly symmetric switching characteristics while all other variations are identical to RPU-baseline model. **(B)** Shows the same training results from (A) at 10x reduced noise levels for all vector-matrix multiplications. **(C)** Pink curve corresponds to the RPU-baseline model trained using "Tiki-Taka" but the noise spec of the forward $A$ cycle in the "Tiki-Taka" algorithm is reduced by 10x. Dark red curve also uses the RPU-baseline model trained using "Tiki-Taka" where the update $C$ cycle of the "Tiki-Taka" algorithm is modified to $C \leftarrow C + \lambda\,(u_t \otimes f(v))$, where $f(v)$ is a pointwise thresholding function that returns $v$ only if $|v| > 0.06$ and otherwise zero. Black, blue lines and orange points\line are plotted again from (A) for comparison.

**Speed, Area and Power Costs**

Compared to the SGD algorithm, the "Tiki-Taka" algorithm introduces additional computations and requires additional hardware resources (crossbar arrays) to perform those computations, and therefore, their area, power and speed costs need to be sized properly.

The "Tiki-Taka" algorithm requires two sets of weight matrices for each layer hence it may increase the area requirement by a factor of 2. In this worst-case scenario $A$ and $C$ matrices can simply be allocated on two separate RPU tiles resulting in twice more area. However, if the RPU devices are integrated at the back-end-of-line (BEOL) in-between metal levels and stacked up as multiple layers, then this area cost can be eliminated. Given that the operations performed on $A$ and $C$ matrices are identical to the ones performed during the SGD algorithm, the same peripheral circuity can be used to drive the lines corresponding to $A$ and $C$ matrices selectively to perform the forward, backward and update cycles in a time multiplex fashion. In this setting, the computations for the forward and backward cycles corresponding to $\gamma = 1$ case can also be realized by driving the lines of $A$ and $C$ matrices simultaneously while integrating the results from both matrices into the same capacitor. Also note that the update cycle on both matrices uses the common stochastic multiplication scheme. Therefore, 4 layers of stacked crossbar arrays can be operated as $A$ and $C$ matrices needed for "Tiki-Taka" without changing the peripheral circuitry design. Given that the same hardware specifications derived in Ref [12] are sufficient for the "Tiki-Taka" algorithm, speed and power of each cycle remains identical to the ones performed in the SGD algorithm. However, "Tiki-Taka" introduced additional cycles to the training and its speed can be easily accounted by simply looking at the $ns$ parameters used during training.

For the FCN-MNIST example $ns = 1$. This setting means that the "Tiki-Taka" algorithm repeatedly performs (1) forward, (2) backward, (3) update, (4) forward and (5) update cycles, 2 additional cycles compared to 3 cycles performed during the SGD algorithm. Since there are not any significant differences between the execution times of the forward, backward and update cycles, the ratio of the wall clock times of "Tiki-Taka" to the SGD algorithm would be 5/3. Increasing $ns$ further decreases this difference as illustrated for the LSTM-WP example where $ns = 5$. In this setting, for every 15 (3 by 5) cycles in the SGD algorithm, the "Tiki-Taka" algorithm introduces 2 additional cycles and hence it runs only ~15% slower than the SGD algorithm. In contrast to the fully connected and LSTM networks where the weight sharing is uniform for all layers, the wall clock time of CNN networks are mainly dictated by the first convolutional layer with the largest weight sharing factor [19]. For the CNN-MNIST example this weight sharing factor is 576. Therefore, even $ns = 1$ is used, for the first convolutional layer the "Tiki-Taka" algorithm introduces 2 additional cycles only after 3x576 cycles. This is a tiny difference and makes the run times of these two algorithms indistinguishable for CNN networks.

We note that there are additional computations that need to be performed outside the crossbar arrays, such as generation of $u_t$ and calculation of $f(v)$. These computations can easily be handled by the digital units that are already responsible for calculating the activations and derivates used in the SGD algorithm. All these additional digital operations performed during the "Tiki-Taka" algorithm are local to the layer and are much simpler than the calculation of activations and derivates, therefore, their relative costs are no more than the relative costs already accounted above for the crossbar arrays.

**Discussion and Summary**

We emphasize that throughout the manuscript we assumed that one crossbar array is used to perform the updates and another separate array is used as fixed reference condunctances. The "Tiki-Taka" algorithm therefore assumes that the updated RPU devices change their conductance bidirectionally and is not directly applicable to one-sided switching devices such as PCM. The stability and convergence of the "Tiki-Taka" algorithm rely on the fact that the random sequence of updates on the $A$ matrix eventually drive all elements of $A$ towards zero. This is indeed achieved by the symmetry point shifting technique, and if this technique is generalized for one-sided switching devices then "Tiki-Taka" can also be used for devices like PCM. However, note that "Tiki-Taka" cannot eliminate the conductance saturation problem. PCM elements change their conductance gradually at one polarity (SET) and very abruptly at the opposite polarity (RESET). Therefore, only SET pulses are send either in the first or the second PCM array depending on the polarity of the weight updates. This eventually results in saturation in the conductance values and therefore require an occasional serial reset operation. None of these complications arise for bidirectional devices and the "Tiki-Taka" algorithm can run with a very limited speed penalty using only parallel operations on the crossbar arrays.

Note that we first derived the hardware induced update rule in presence of asymmetric devices and then showed its relevance to the SGD algorithm. For instance, for some specific device switching characteristics the hardware induced update rule looks similar to adding $\ell_2$ regularization term into the optimization objective. However, the strength of this additional term is large and not controllable, and hence resulting in poor training results. Note for the $\gamma = 0$ case of "Tiki-Taka" the weights of the neural network are stored in $C$ which are updated using the gradients accumulated on $A$. In this setting the hardware induced rules on $A$ and $C$ matrices show resemblance to the momentum-based SGD algorithm providing further intuition to the "Tiki-Taka" algorithm. However, careful investigation shows that the "Tiki-Taka" algorithm is not just an instance of the momentum-based SGD and may require further investigation. One obvious direction is to test the "Tiki-Taka" algorithm for networks even larger than the ones presented here.

In summary, we proposed a new DNN training algorithm, so called "Tiki-Taka" algorithm, that uses a coupled system in order to simultaneously minimize the objective function of the original network of interest and the hidden cost term that is unintentionally introduced due to asymmetric device switching characteristics. Training simulations performed on various network architectures show that even a very aggressive device asymmetry can be compensated by "Tiki-Taka" giving indistinguishable training results compared to ones achieved with the perfectly symmetric (ideal) devices. We emphasize that the asymmetry behavior used in our simulations and shown to be sufficient for "Tiki-Taka" is already experimentally observed by many device technologies but declared unsatisfactory due to asymmetry. Assuming other device specifications are within the tolerable margins, all those nonsymmetric device technologies can now be used for deep learning applications, as the "Tiki-Taka" algorithm significantly relaxes the challenging symmetric switching criteria needed from the resistive cross-point devices.